\documentclass{article}

\usepackage{PRIMEarxiv}

\usepackage[utf8]{inputenc} 
\usepackage[T1]{fontenc}    
\usepackage{hyperref}       
\usepackage{url}            
\usepackage{booktabs}       
\usepackage{amsfonts}       
\usepackage{nicefrac}       
\usepackage{microtype}      
\usepackage{fancyhdr}       
\usepackage{graphicx}       
\graphicspath{{media/}}     
\usepackage{cite}
\usepackage{amsmath, amssymb}
\usepackage{url} 
\usepackage[dvipsnames]{xcolor}

\usepackage{makecell}
\usepackage{pifont}
\usepackage{subcaption}

\newcommand{\secref}[1]{Section~\ref{#1}}
\newcommand{\psecref}[1]{(see \secref{#1})}
\newcommand{\figref}[1]{Figure~\ref{#1}}
\newcommand{\capref}[1]{Chapter~\ref{#1}}
\newcommand{\tabref}[1]{Table~\ref{#1}}

\newcommand{\red}[1]{\textcolor{Red}{#1}}
\newcommand{\green}[1]{\textcolor{Green}{#1}}
\newcommand{\blue}[1]{\textcolor{Cyan}{#1}}

\newcommand{\ffigure}[5]{
    \begin{figure}[ht]
        \centering
        \includegraphics[width={#5},height={#4}]{#2}
        \caption{#3}
        \label{#1}
    \end{figure}
}

\DeclareMathOperator{\Conv}{Conv}
\DeclareMathOperator{\PReLU}{PReLU}
\DeclareMathOperator{\Inorm}{IN2d}
\DeclareMathOperator{\Sigmoid}{Sigmoid}
\DeclareMathOperator{\Mix}{Mix}
\DeclareMathOperator{\Up}{Up}

\newcommand{\cmark}{\ding{55}}%
\newcommand{\xmark}{\ding{51}}%

\title{
  TinyCD: a (not so) Deep Learning Model for Change Detection
}

\author{
  Andrea Codegoni\\
  Dipartimento di Matematica "F. Casorati"\\
  University of Pavia \\
  \texttt{andrea.codegoni01@ateneopv.it} \\
   \And
  Gabriele Lombardi, Alessandro Ferrari \\
  ARGO Vision \\
  Milano\\
  \texttt{\{gabriele.lombardi,alessandro.ferrari\}@argo.vision} \\
}

\begin{document}
\maketitle

\begin{abstract}
In this paper, we present a lightweight and effective change detection model, called TinyCD.
This model has been designed to be faster and smaller than current state-of-the-art change detection models due to industrial needs. 
Despite being from 13 to 140 times smaller than the compared change detection models, and exposing at least a quarter of the computational complexity,
our model outperforms the current state-of-the-art models by at least $1\%$ on both F1 score and IoU on the LEVIR-CD dataset, 
and more than $8\%$ on the WHU-CD dataset.
To reach these results, TinyCD uses a Siamese U-Net architecture exploiting low-level features in a globally temporal and locally spatial way.
In addition, it adopts a new strategy to mix features in the space-time domain 
both to merge the embeddings obtained from the Siamese backbones, 
and, coupled with an MLP block, it forms a novel space-semantic attention mechanism, the Mix and Attention Mask Block (MAMB).
Source code, models and results are available here: \url{https://github.com/AndreaCodegoni/Tiny_model_4_CD}
\end{abstract}


\section{Introduction}\label{sec:intro}

In the Remote Sensing community, Change Detection (from now denoted with CD) is one of the main research topics.
The main purpose of CD is to identify changes occurred in a scene between two different times.
To this aim, a CD model compares two co-registered images $I_1$ and $I_2$
acquired at times $t_1$ and $t_2$ \cite{singh1989review,shafique2022deep,bai2022deep}.
Once the relevant changes have been identified, such as urban expansion, deforestation, or post disaster damage assessment 
\cite{chen2020spatial,de2020change,vina2004satellite,xu2019building,zhang2020deeply,ji2018fully,varghese2018changenet},
the challenge is to let the CD model ignore other irrelevant changes.
Examples of irrelevant changes are, but not limited to, lighting conditions, shadows, and seasonal variations.

Thanks to the increasing number of available high resolution aerial images datasets,
such as \cite{chen2020spatial,ji2018fully,zhang2020deeply}, 
data driven methods like deep Convolutional Neural Networks (CNN) found successful applicability \cite{khelifi2020deep}.
The well known ability of deep CNNs to extract complex and relevant features from images 
is the key factor for their early promising results \cite{daudt2018fully}.
In the CD scenario, complex features are important, but are not sufficient to accomplish the task.
To detect the occurred changes, it is in fact crucial to model the spatio-temporal dependencies between the two images. 
Unfortunately, plain CNNs have a limited receptive field due to the usage of fixed kernels in convolutions.
To overcome this issue, recent works focused their attention to 
enlarging the receptive fields by employing different kernel types \cite{zhang2018triplet},
or by adding attention mechanisms 
\cite{liu2020building,zhang2020deeply,peng2019end,peng2019end,jiang2020pga,chen2020spatial,chen2020dasnet}. 
However, most of them failed to explicitly relate data in the temporal domain, since attention mechanisms are applied 
separately on the two images. 
The self attention mechanism adopted in \cite{chen2020spatial,chen2020dasnet}
shows promising results relating images in the spatio-temporal domain. 
More recently, Transformers have been introduced in CD because of 
their receptive fields spatially covering the whole image \cite{chen2021remote,bandara2022transformer}. 
Notice that, by applying multi-headed attention layers in the decoder part of the network,
the receptive field covers the temporal domain too.
Unfortunately, the resulting models are computationally very inefficient.

The CD field finds applicability also outside the remote sensing world.
As an example, in \cite{varghese2018changenet,chen2021dr} two models are discussed in order to be used on drones 
or other autonomous vehicles to implement smart city monitoring functions. 
In our case, the change detection model has been developed for an industrial application.
In our application field, the need for real-time performances adds a model complexity constraint.
Unfortunately, the majority of state-of-the-art (SOTA) models are millions-parameters-sized, so that their applicability is not possible.
Another issue with those big models is related to the training time clearly affected by the size of the model. 
With large models, the Hyper-Parameters-Optimization (HPO) task requires resources that are usually not available to medium-small companies.
Moreover, big networks require dedicated hardware also at inference time.
This is in contrast with production requirements and project budgets.
The search for models having both small size and performances comparable to the current SOTA can be considered an open problem.

A possible strategy to cope with both model size and complexity that, to the best of our knowledge, 
has not been studied in the literature, is to use low-level features to compare the two images under examination.
Another underestimated aspect, in our opinion, is that a Siamese type backbone produces two tensors 
containing channels arranged semantically in the same order.
This observation could be used to design strategies for merging features more efficiently.

The main purpose of our work is to investigate the aforementioned issues developing a neural network 
that requires lower computational complexity with respect to the SOTA CD models reaching at the same time comparable performances.

The major contributions of our work are the following:
\begin{itemize}
    \item We explore the effectiveness of using low-level features in the problem of comparing images. 
          The results validate our intuition that in this context the low-level features are sufficiently expressive. 
          Moreover, this allowed us to significantly limit the number of model parameters.

    \item We introduce a novel strategy to mix the features between the two images. 
          This strategy allows the computation of a spatio-temporal correlation between the input images
          keeping a low computational complexity.

    \item A fast attention mechanism is introduced with a block called MAMB.
          It uses features localized in space to compute attention masks needed in the 
          up-sampling phase to refine the low-resolution results. 

    \item We propose to use a pixel-wise classifier to generate the final mask.
          In our tests, this proved to be very effective.
\end{itemize}

Our architecture exploits the information contained in the channels of the feature vectors generated by the backbone.
For this reason, it can effectively exploit low level features such that a relatively small backbone can be adopted.
Being the backbone%
\footnote{Notice that the backbone is evaluated twice in Siamese architectures.} 
the most time-consuming and parameter-demanding component in the architecture,
maintaining it as small as possible allows us to achieve our goal.
In particular, this allows us to maintain the total number of parameters below $300000$.

Finally, we compare the quality of the model with SOTA architectures, and we demonstrate that
it has performances comparable if not even superior to other SOTA models in the CD field.
We have extensively tested our model on public and proprietary datasets. 
In order to validate and make reproducible our results,
in this paper we highlight the results obtained in the field of aerial images on public datasets. 
Similar results in terms of efficiency and effectiveness have been found in non-public datasets, 
in application fields other than the one faced in this paper.

The paper is organized as follows. In \capref{cap:relatedworks} we present some related works.
In \capref{cap:proposedmodel} we describe our proposed model. In \capref{cap:results} we 
report both the results of our model on two publicly available datasets and the ablation study. 
Finally, in \capref{cap:conclusion} we highlight some future research directions. 

\section{Related works} \label{cap:relatedworks}
\subsection{Early deep neural network works on CD}
Deep learning models, and in particular CNNs, 
have been applied with great success in image comparison tasks
\cite{chopra2005learning,zagoruyko2015learning,stent2015detecting}, 
in pixel-level image classification \cite{long2015fully,ronneberger2015u,bertinetto2016fully}, 
and they represent the SOTA in many other Computer Vision fields \cite{ioannidou2017deep}. 

Models in the context of the CD must manage two inputs: 
one image $I_1$ acquired at time $t_1$, and another one $I_2$ acquired at time $t_2$.
The correct use of these two inputs, and the features extracted from them,
are extremely important for the well behavior of the CD model.
One of the first works using deep learning techniques to the field of CD is \cite{chu2016change}. 
This work highlights how deep neural networks, 
in particular Deep Belief Networks obtained by stratifying Restricted Boltzmann machines, 
are a very valid tool to compare and highlight the changes between the two images under examination.
To the best of our knowledge, the first work that applies CNNs to the CD problem is \cite{daudt2018fully}. 
In this work the authors propose two different approaches.
In the first case they use a 
U-Net \cite{ronneberger2015u} type network with the Early Fusion Strategy (FC-EF), 
i.e. they concatenate the images $I_1$ and $I_2$,
and then they feed the U-Net with the resulting tensor. 
In the second case
they investigate the Feature Fusion Strategy.
To this aim, they employ a Siamese U-Net type network \cite{bromley1993signature,zagoruyko2015learning,bertinetto2016fully} 
where the two images are processed separately, and subsequently the features are fused in two different ways: 
concatenation (FC-Siam-conc) and subtraction (FC-Siam-diff).
These fused features are then used as skip connections in the decoder.
After this seminal work, an entire research line investigated both the Early Fusion Strategy 
\cite{de2020change,lebedev2018change,peng2019end,zhao2020using},
and the Feature Fusion Strategy  
\cite{chen2020spatial,chen2020dasnet,zhang2018triplet,liu2020building,
zhang2020deeply,peng2020optical,jiang2020pga,bao2020ppcnet,hou2019w,zhan2017change,fang2019dual,chen2021adversarial}.

To take full advantage of the large amount of spatial information, deeper CNNs such as ResNet \cite{he2016deep} 
or VGG16 \cite{simonyan2014very} have been used 
\cite{chen2020spatial,chen2020dasnet,zhang2018triplet,zhang2020deeply} in order to extract spatial information 
and group them in a hierarchical way.
Unfortunately, standard convolution has a fixed receptive field that limits the capacity
of modelling the context of the image.
To face this issue, atrous convolutions \cite{chen2017deeplab} have been experimented \cite{zhang2018triplet}:
they are able to enlarge the receptive field of convolutional kernels
without increasing the number of parameters.

\subsection{Attention based CNN}
To definitively overcome the problem of fixed receptive field,
attention mechanisms, in the forms of spatial attention \cite{liu2020building,zhang2020deeply,peng2019end}, 
channel wise attention \cite{liu2020building,zhang2020deeply,peng2019end,jiang2020pga}, 
and also self-attention \cite{chen2020spatial,chen2020dasnet}, have been introduced.
In \cite{zhang2020deeply}, the attention mechanisms are used in the decoder part: 
the channel wise attention is used to re-weight each pixel after the fusion with the skip connections,
while the spatial attention is adopted to spatially re-weight the pixels containing misleading information due to the up sampling step.
To further exploit the interconnection between spatial and channel information,
in \cite{liu2020building} a dual attention module has been introduced. 
The co-attention module introduced in \cite{jiang2020pga} tries to leverage correlation between features
extracted from both images. Also in \cite{jiang2020pga} a co-layer aggregation and a pyramid structure is used 
to make full use of the features extracted at each level and with different receptive fields. 
In \cite{chen2020spatial}, the non-local self attention introduced in \cite{wang2018non},
have been applied to CD. 
This mechanism consists in stacking the features extracted from a Siamese backbone
to then apply both a basic spatial attention mechanism, and a pyramidal attention mechanism.
Since these two attention blocks are applied to stacked features obtained from $I_1$ and $I_2$,
these are correlated in a non-local spatio-temporal way.
Another interesting approach is the one presented in \cite{liu2021change}. In this paper, 
the authors decided to combine CNNs with Object Image Analysis (OBIA) to mitigate the limited receptive field problem. 
In a first preprocessing phase, they segment the image and extract the patches containing objects to be compared. 
Subsequently, the extracted patches are compared using a CNN 
which then works on small patches containing more specific and detailed information.

In \cite{chen2021dr}, the authors propose a temporal attention mechanism. 
They exploit the features extracted from $I_2$ to generate a query matrix 
which is then compared with the features extracted from $I_1$.
This mechanism is made dynamic by reducing the receptive field as tensors' spatial dimension diminishes.
Finally, the authors also use attention mechanisms capable of emphasizing some horizontal and vertical dependencies 
of recurring objects in their scenario.

\subsection{Transformers in CD}
\label{sec:traformersincd}
Finally, the global attention mechanisms introduced with Transformers 
\cite{wu2020visual,vaswani2017attention} have also been applied to the CD problem. 
In \cite{chen2021remote} the authors employ a modified ResNet18 as Siamese backbone to extract features. 
Then, to better justify the use of Transformer blocks,
they follow a parallelism between the natural language processing field,
and the image processing one, by introducing the semantic tokens.
Roughly speaking, semantic tokens are the pixels of the last feature tensor extracted by the backbone.
The authors use this concept to illustrate that concatenating single pixels and then processing them with a transformer 
encoder-decoder, a pair of features tensors can be obtained that incorporates both global spatial information, and global temporal information. 
On the other side, in \cite{bandara2022transformer} the authors replace the CNN backbone with a transformer in order to exploit the global 
information contained in the images right from the start. 
In this model, the temporal aggregation is done only in the final multilayer perceptron decoder.

\subsection{Relations between our work and existing models}
\label{chp:relationexisting}
Our work is inspired by \cite{chen2021remote}. As reported in Section \ref{sec:traformersincd}, in that work the authors
introduce the concept of semantic tokens, which are basically single pixels of the tensor obtained by the backbone.
Then, they use a transformer in order to process these tokens and extract global spatio-temporal information.
In agreement with \cite{chen2021remote}, we believe that the information contained in the 
pixel/semantic token is crucial to obtain a good result.
However, we prefer to apply channel-wise local feature comparison,
limiting the semantic complexity and aggregation of adopted features to the first few backbone layers;
whilst in \cite{chen2021remote} the comparison is global, being it obtained by means of transformers.
Moreover, we adopt Multi Layer Perceptrons (MLPs) to compute both the spatial attention maps and the final mask,
actually facing the problem as a pixel-wise classification one.
Recently, MLP blocks have received great attention in computer vision 
\cite{chen2021cyclemlp,lian2021mlp,zhang2022bending,tolstikhin2021mlp,touvron2022resmlp,liu2021pay,yu2022s2}.
These architectures divide the images into patches and then process the patches with MLP blocks.
Different structures of MLP blocks have been proposed to incorporate as much spatial information as possible.
For example, in \cite{yu2022s2,lian2021mlp} a spatial shift operator is applied in order to 
obtain information from different axial directions. 
The CycleMLP block proposed in \cite{chen2021cyclemlp} follows a similar idea 
but instead of applying the spatial shift operator to the features' tensor,
it composes several MLP steps capable of mimic the shift.
A more refined version of these concepts is proposed in \cite{zhang2022bending} were the authors employ a block 
which dynamically learns the spatial offset used in CycleMLP.
In our model, the MLP blocks work exclusively along the channel dimension
to both compute the spatio-temporal attention maps, and produce the final pixel-wise classification.

\section{Proposed model}\label{cap:proposedmodel}

In this section we describe and motivate the structure of our model.
We use a model resembling a Siamese U-Net consisting of 4 main components:
\begin{itemize}
    \item Siamese encoders constituted by a pre-trained backbone \psecref{sec:backbone}.
    \item Mix and Attention Mask Block (MAMB) and bottleneck mixing block to compose backbone results \psecref{sec:MAMB}.
    \item Up-sample decoder to refine low resolution results incorporating higher resolution data from the skip connections \psecref{sec:decoder}.
    \item Pixel level classifier \psecref{sec:classifier}
\end{itemize}

\ffigure{fig:whole_model}{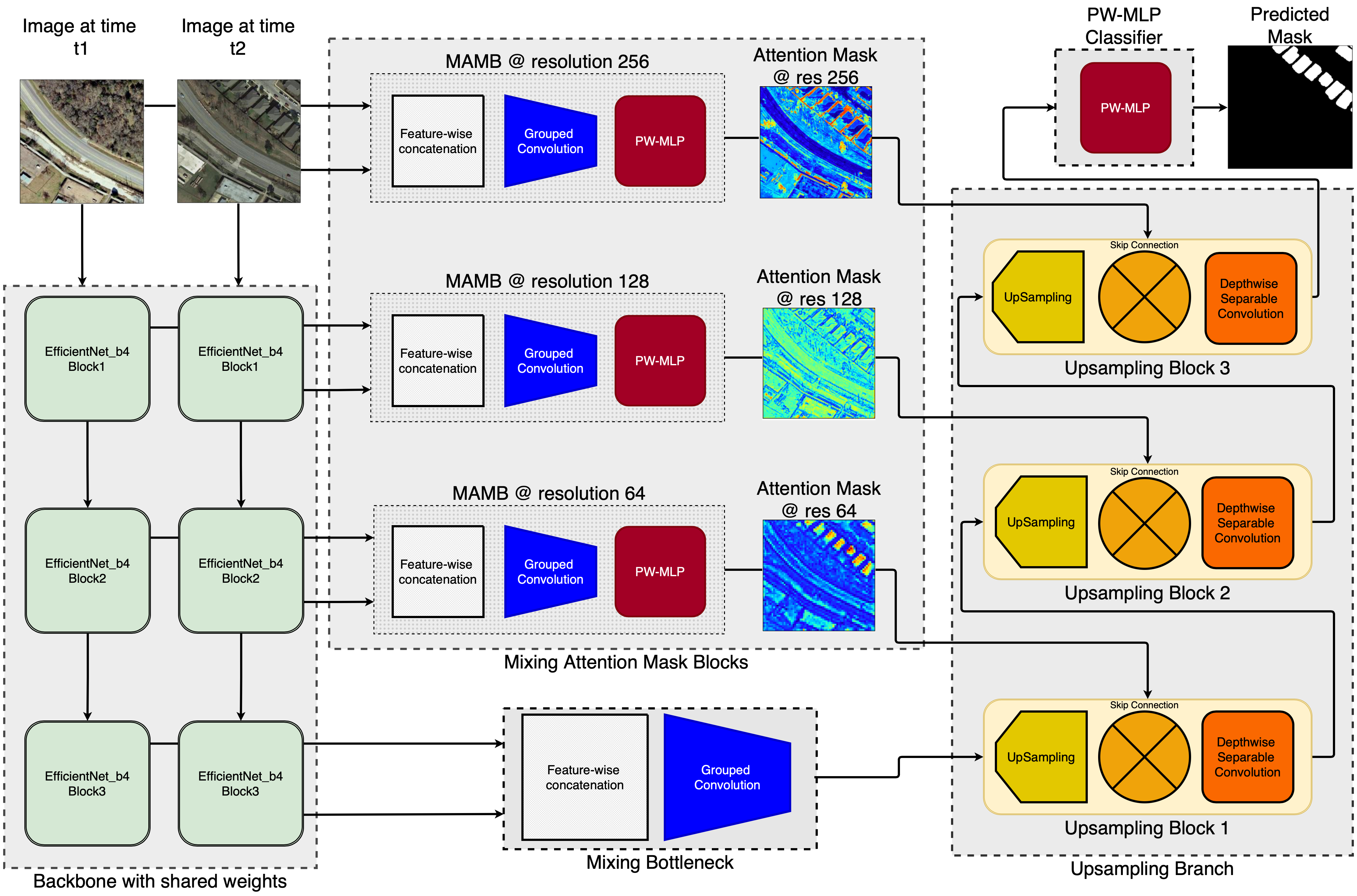}{
   Siamese U-Net architecture including MAMB.
}{7.5cm}{12cm}

In what follows, we denote with $X\in\mathbb{R}^{(C \times H \times W)}$ the \emph{reference} tensor (image at time $t_1$) 
and with $Y\in \mathbb{R}^{(C \times H \times W)}$ the \emph{comparison} tensor (image at time $t_2$). 
$C$ is the number of channels, $H$ is the height and $W$ the width of the tensors. We omit the batch dimension for the ease of notation. 
We denote with $\Conv$ the convolution operator, with $\PReLU$ the Parametric Rectified Linear Unit \cite{he2015delving}, 
with $\Inorm$ the Instance Normalization \cite{ulyanov2016instance}, and $\Sigmoid$.

\subsection{Model overview}

Indicating with $f_k$ the composition of the backbone blocks up to the $k^{th}$ one,
the high-level features $X_k=f_k(X)$ and $Y_k=f_k(Y)$ are extracted from each level $k$ of the backbone.
These features are used both to compute the resulting output at each level of the U-Net encoder,
and to estimate the attention masks. 
The last backbone block produces the embeddings $X_e$ and $Y_e$ representing the bottleneck inputs.

Every backbone intermediate output pair $(X_k,Y_k)$ is processed by means of the MAMB,
producing spatial attention masks $M_k$.
These masks are used as skip-connections and composed in the decoder.
The last mixed tensor is obtained by composing $(X_e,Y_e)$.

The decoder consists of a series of up-layers, one for each block of the backbone. 
Each up-layer increases the spatial dimensions of the tensor received from the previous layer
to reach the same resolution of the corresponding skip-connection.
Furthermore, the up sampled tensors and the skip-connections are composed to generate the next layer inputs.
This composition is the attention mask application to the features obtained from the previous layer.

Finally, the last block of our model classifies each pixel of the obtained tensor through a Pixel-Wise Multi-Layer Perceptron (PW-MLP).
The PW-MLP associates to each pixel the probability that it belongs to the anomaly class.
Applying a threshold to this tensor we obtain the binary mask of changes.

In the following subsections we describe each component separately.

\subsection{Siamese encoders with pre-trained backbone}
\label{sec:backbone}

The purpose of the Siamese encoder is to extract simultaneously features from both images in a semantic coherent way.
In deep neural networks, training the first layers of the model is sometimes difficult due to the well-known phenomenon of 
vanishing gradients \cite{hochreiter2001gradient,hochreiter1997long}.
To overcome this problem, several tricks have been introduced such as the residual connections of 
ResNet \cite{he2016deep}, or the skip connections of the U-Net \cite{ronneberger2015u}. 
However, training deep backbones remains a difficult, time-consuming, 
or even impossible task to accomplish if the dataset is too small. 

For these reasons, pre-trained backbones are often preferred, 
even in CD problems \cite{chen2020spatial,chen2020dasnet,zhang2018triplet,zhang2020deeply,chen2021remote}.
The disadvantage of this approach is that the backbones are not always trained on images that are similar to the ones we are dealing with.
However, CNN backbones work by layering information. Low-level features, such as lines, 
black/white spots, points, edges, can be considered general-purpose being common to all images.

In our intuition, in the faced task the comparison between two images 
$I_1$ and $I_2$,
can be accomplished by using just the low-level features extracted from the first few layers of a pre-trained backbone.

We therefore decided to use one EfficientNet backbone \cite{tan2019efficientnet} pre-trained on the ImageNet dataset \cite{deng2009imagenet}.
We allowed the training phase to tune also the totality of the backbone parameters.
Guided by experiments on our industrial dataset, 
the EfficientNet backbone family have been selected due to both its efficacy and its efficiency.
Moreover, the resolution reduction in the first EfficientNet layers is sufficiently slow 
in order to create skip connections of different spatial dimensions.

For completeness, in Appendix \ref{appx:backbone} 
we compare the performance of other backbones 
in order to show the generality of our approach.

\subsection{Mix and Attention Mask Block (MAMB) and bottleneck mixing block}
\label{sec:MAMB}

The purpose of this block is to merge the features $(X_k,Y_k)$ extracted from one of the blocks of the Siamese encoder.
It creates a mask $M_k$ that is then used as skip connection to refine the information obtained during the up-sampling phase.

The mask we create can also be understood as a pixel-level attention mechanism.
The idea of pixel-wise attention has been already studied in \cite{liu2018picanet}.
Here we specifically designed a pixel-wise attention mechanism exploiting both spatial and temporal information.

The MAMB can be divided into two sub-blocks: 
the Mixing block \psecref{sec:mixing},
and the Pixel level mask generator \psecref{sec:pixel-mask}.

\subsubsection{Mixing block}
\label{sec:mixing}

As the name suggests, in this sub-block we compose the features generated by the $k^{th}$ backbone blocks $(X_k,Y_k)$.
To this aim, we observe that the features $X_k$ and $Y_k$, share both the same shape $C_k,\ H_k,\ W_k$, and 
the same arrangement in terms of features. 
This means that the features in channel $c$ of $X_k$ have the same semantic meaning with respect to the corresponding 
features in channel $c$ of $Y_k$, being the Siamese encoder weights shared.
In view of this observation, we decided to concatenate the tensors $X_k$ and $Y_k$ in the tensor 
$Z_k \in \mathbb{R}^ {2C_k \times H_k \times W_k}$
using the following rule:
\begin{equation}
      Z_k^c := 
      \begin{cases}
         X_k^{c/2}     & \text{$c$ even} \\
         Y_k^{(c-1)/2} & \text{$c$ odd} \\
      \end{cases}
      \quad\forall c \in \{0,...,2C_k-1\}.
   \label{eq:mixtensor}
\end{equation}

To mix the features coming from  $X_k$ and $Y_k$ both spatially and temporally, we used a group convolution.
By choosing the number of groups equal to $C_k$ we obtain $C_k$ kernels of depth 2 which process the tensor $Z_k$ in pairs of channels. 
These kernels perform at the same time both spatial and temporal convolution
using the cross-correlation between semantically similar features.

The new tensor $Z_k' \in \mathbb{R}^{C_k \times H_k \times W_k}$ is defined as:
\begin{align}
\begin{split}
Z_k' = \Mix(X_k,Y_k) := \PReLU [&\Inorm [\Conv(Z_k,\ ch_{in} = 2C_k, \\
& ch_{out} = C_k,\ groups = C_k) ]].
\label{eq:convmix}
\end{split}
\end{align}

An illustration of our concatenation strategy, and the following grouped convolution, is reported in \figref{fig:mix_conv}.

\ffigure{fig:mix_conv}{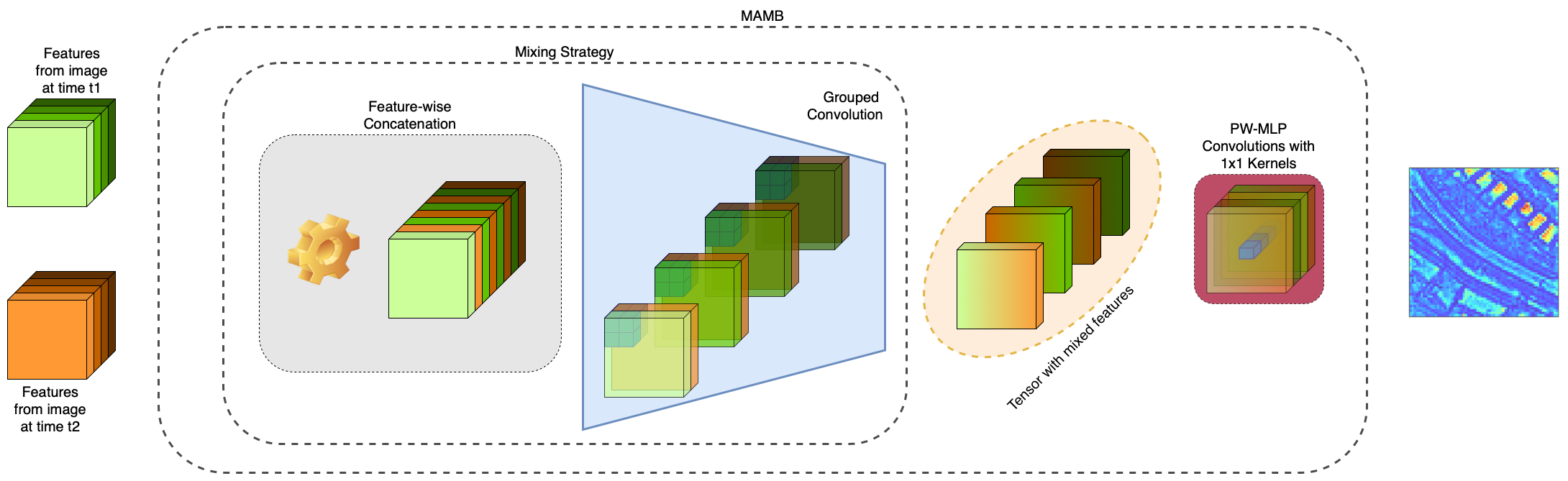}{
   Visual representation of our mixing strategy and the full MAMB block.
   In the inner dashed block we highlight the
   concatenation strategy, \eqref{eq:mixtensor}, the grouped convolution, \eqref{eq:convmix}.
   These two blocks when coupled with the PW-MLP form the MAMB block.
}{3.5cm}{10cm}

\subsubsection{Pixel-level mask generator}
\label{sec:pixel-mask}

Fixing the spatial coordinates of a single pixel, the $C_k$ values in the tensor $Z_k'$ 
contain spatial information related to both times $t_1$ and $t_2$.
Our idea is to use the PW-MLP in order to process this information and generate a score that acts as a spatio-temporal attention. 

To this aim, the PW-MLP is designed to produce a mask tensor $M_k\in \mathbb{R}^{H \times W}$.

\subsubsection{PW-MLP}
\label{sec:PW-MLP}

To implement a pixel-wise Multi-Layer Perceptron,
that is an MLP working on all the channels of one single pixel at a time,
we use $1 \times 1$ convolutions.
The MLP is composed by $N$ blocks each containing one $1 \times 1$ convolution and one activation function.
As activation, we used the $\PReLU$, being this able to propagate gradients also on the negative side of the real axis.
The last convolution contains just one filter, 
thus producing a tensor $M_k$ with dimensions $1,\ H_k,\ W_k$.

The use of $1 \times 1$ convolutions to implement an MLP is not a new idea.
In \cite{lin2013network} this strategy has been used to substitute layers such as convolutions with small, trainable, networks.
As pointed out in \cite{lin2013network}, we have very poor prior information on the latent concepts in pixel vectors.
Hence, we have decided to use this universal function approximator to separate different semantic concepts.

\subsubsection{The bottleneck mixing block}
\label{sec:bottleneck}

We applied the tensor mixing strategy reported in \secref{sec:mixing}
to compute the bottleneck of the U-Net like network.
More precisely, we compute: $U_e = \Mix(X_e,Y_e)$.

$U_e$ represents the output of the encoder and the input to be processed by the decoder.
$U_e$ contains the spatially and temporarily correlated higher level features computed by the backbone.
Given that, our intuition is that $U_e$ contains enough information in order to classify each pixel at the bottleneck resolution.

\subsection{Up-sampling decoder with skip connections}
\label{sec:decoder}

The general $k-$th decoder block takes as input the tensor $U_{k+1}$ of shape $C_{k+1}, H_{k+1}, W_{k+1}$ and 
a mask $M_{k}$ of shape $1, H_{k}, W_{k}$.
Firstly, an up sampling operation is performed in order to transform $U_{k+1}$ so that its shape matches the one of $M_k$.
We call the up sampled tensor $U_{k}'$.
Then, we define $U_{k}$ with
\begin{equation*}
   U_k := \PReLU\left[\Inorm\left[\Conv(U_k' \odot M_k)\right]\right],
\end{equation*}
where we have denoted with the symbol $\odot$ the Hadamard product.
This represents the skip connection attention mechanism at the pixel level. 

As we already mentioned in \secref{sec:bottleneck}, $U_e$ contains enough information to classify each pixel at its spatial resolution.
By multiplying the mask $M_k$, we are re-weighting each pixel in order to alleviate the misleading information generated by up sampling.

Notice that, in this $\Up$ block we employ the depth wise separable convolution \cite{sifre2014rigid,chollet2017xception}.

\subsection{Pixel level classifier}
\label{sec:classifier}

Finally, since the change detection problem is a binary classification problem,
we decided to use as last layer a PW-MLP with output classes $\{0,1\}$ representing respectively normal and changed pixels.
With respect to what reported in \secref{sec:PW-MLP},
in this case we used as the last activation layer a $\Sigmoid$ function instead of the $\PReLU$, 
thus enforcing the result of the network to contain values in $[0,1]$.
In this case, the PW-MLP is used as a non-linear classifier which
separates pixels in normal or changed class. 

\section{Experiment Settings and Results} \label{cap:results}

In this section we presents the settings used in our experiments, the achieved results, and the performed ablation study.

\subsection{Datasets}

As already stated in \secref{sec:intro}, we cannot share the dataset related to our industrial application.
Moreover, in order to fairly evaluate our model, and to compare it with other works in the CD field,
we used the following public and widely adopted aerial building images datasets: LEVIR-CD \cite{chen2020spatial} and WHU-CD \cite{ji2018fully}%
\footnote{
    Both the adopted datasets have been obtained from \url{https://github.com/wgcban/SemiCD} in an already pre-processed version.
}. 
Notice that the task defined by these datasets is particularly close to the faced industrial one, that is the driver of our research work.
In these two datasets the model has to track some specific patters, those corresponding to buildings, and carefully segments the eventually occurred changes.

LEVIR-CD contains 637 pairs of high resolution aerial images. 
Starting from these images, patch pairs of size $256\times256$ each have been extracted.
After that, the pair instances have been partitioned accordingly to the authors' original indications.
This step produced 7120, 1024, and 2048 pair instances for the train, validation, and test dataset respectively.

WHU-CD contains just one pair of images having resolution $32507\times 15354$ as a crop of a wider geographic area%
\footnote{
    The whole dataset depicts the city of Christchurch, in New Zealand. 
    The crop, aimed to be used in CD tasks, is a sub-area acquired in two different times.
}.
Following \cite{bandara2022revisiting}, the images have been split in non overlapping patches with resolution $256 \times 256$.
After that, a randomly partitioning of the dataset have been performed obtaining 5947, 743, and 744 pairs for train, validation, and test respectively.

\subsection{Loss function and evaluation metrics}

As stated in \secref{sec:classifier}, we cast the CD problem in a pixel-wise binary classification setting.
In fact, the role of the final MLP block is to output the per-pixel change probability.

Since the reference mask is a binary mask (0 for unchanged pixels, 1 for changed pixels), and since we are comparing 
probabilities, one loss function that can be used is the Binary Cross Entropy (BCE). It is defined as:
\begin{align*}
\mathcal{L}(G,P):=
    -\frac{1}{|H|\cdot|W|}&\sum_{h\in H,w\in W} g_{h,w}\log(p_{h,w}) +\\
    &(1-g_{h,w})\log(1-p_{h,w}),
\end{align*}

where we denoted with $G$ the ground truth mask, with $P$ the model prediction, and with $H$ and $W$ the set of indices relative to height and width.

Notice that the BCE loss function is widely used in other SOTA models such as \cite{bandara2022transformer,chen2021remote}.
In contrast, other researchers implemented more sophisticated loss functions like the one presented in \cite{chen2020spatial}.
We decided to use the simpler BCE in order to attribute the improvement in performances to the model and not to an ad hoc built-in loss function.
For completeness, in appendix \ref{appx:A1} we report other experiments conducted using other widely adopted loss functions.

To evaluate the performances achieved by our model, we calculated the
\emph{Precision (PR)}, \emph{Recall (RC)}, \emph{F1 score (F1)}, \emph{Intersection over Union (IoU)}
and \emph{Overall Accuracy (OA)} with respect to the change class, as defined below:
\begin{align*}
    &Pr := \frac{TP}{TP+FP}, \\
    &Rc := \frac{TP}{TP+FN},\\
    &F1 := \frac{1}{Pr^{-1}+Rc^{-1}}, \\
    &IoU := \frac{TP}{FN+FP+TP},\\
    &OA := \frac{TP+TN}{FN+FP+TP+TN},
\end{align*}
where $TP$, $TN$, $FP$, $FN$ are computed on the change class, and represent the true positives, true negatives, false positives, and false negatives respectively.
To retrieve the change mask we applied a $0.5$ threshold to the output mask.

\subsection{Implementation details}

We implemented our model using PyTorch~\cite{paszke2019pytorch} and we trained it on an NVIDIA GeForce RTX 2060 6GB GPU.
As described in \secref{sec:backbone}, we selected the first four blocks of the EfficientNet version $b4$ backbone pretrained on the ImageNet dataset.
All other weights of the model have been initialized randomly%
\footnote{
    To make our results reproducible, we fixed the random seed at the beginning of each experiment.
}.

As optimizer, we adopted AdamW \cite{loshchilov2017decoupled}.
To optimize its hyperparameters, i.e. learning rate, weight decay and \emph{amsgrad} variant, 
and also to verify the robustness of our model with respect to the choice of these parameters,
we firstly run a Hyper-Parameters Optimization (HPO) task for each dataset using the package Neural Network Intelligence (NNI) \cite{nni2021}.
After this, we fixed the learning rate equal to $3\cdot10^{-3}$, and the weight decay equal to $9\cdot10^{-3}$, for the LEVIR-CD dataset.
Moreover, we fixed the learning rate equal to $2\cdot10^{-3}$, and the weight decay equal to $8\cdot10^{-3}$, for the WHU-CD dataset.
For both datasets, \emph{amsgrad} have been set to \emph{False}.
An example of the HPO procedure is reported in Appendix \ref{appx:A1}.
Due to computational resource limitations, no other hyperparameters have been tuned.
We have not experimented network architecture search techniques (NAS).

To dynamically adjust the learning rate during the training, 
we adopted the cosine annealing strategy as described in \cite{loshchilov2016sgdr},
but avoiding the warm restart.

Since aerial images are spatially registered, we applied the geometric data augmentation operators 
simultaneously to the reference/comparison images and their associated ground-truth mask.
Also, non-geometric augmentations are applied independently on the reference and the comparison images.

The applied geometric augmentations are Random Flip on both X and Y axes, and Random Rotation with free degree.
Moreover, the applied non-geometric augmentations are Gaussian Blur and Random Brightness/Contrast change.
To achieve all the adopted augmentations, we used the Albumentations library \cite{buslaev2020albumentations}.

Finally, due to the limited GPU memory capacity and computational power, we fixed the batch size to 8, and trained for just 100 epochs.

\subsection{Comparison with SOTA models}

To demonstrate the effectiveness of our approach, we compared our results with those reported in \cite{chen2021remote,bandara2022transformer}. 
As baseline, we used the three models presented in \cite{daudt2018fully}. 
Moreover, to compare our model with other works adopting both spatial and channel attention mechanisms,
we dealt with \cite{liu2020building,chen2020spatial,zhang2020deeply,fang2019dual}. 
Finally, given the success achieved by Transformers applied to the computer vision field, 
we also compared our results with those obtained in \cite{chen2021remote,bandara2022transformer}. 

The results reported in \tabref{tab:metrics-levir} and \tabref{tab:metrics-whu} show the superior performance of our model 
on the LEVIR-CD and WHU-CD building change detection datasets. 

\begin{table}[ht]
    \caption{
        Performance metrics on the LEVIR-CD dataset.
        To improve results readability, we adopted a color ranking convention to represent the \green{First}, \red{Second}, and \blue{Third} results respectively.
        The metrics are reported in percentage.
    }
    \centering
    \begin{tabular}{l|ccccc}
        \multicolumn{6}{c}{LEVIR-CD} \\
        \hline 
        Model & Pr & Rc & F1 & IoU & OA \\
        \hline 
        FC-EF \cite{daudt2018fully}                 & 86.91 & 80.17 & 83.40 & 71.53 & 98.39 \\
        FC-Siam-diff \cite{daudt2018fully}          & 89.53 & 83.31 & 86.31 & 75.92 & 98.67 \\
        FC-Siam-conc \cite{daudt2018fully}          & 91.99 & 76.77 & 83.69 & 71.96 & 98.49 \\
        DTCDSCN \cite{liu2020building}              & 88.53 & 86.83 & 87.67 & 78.05 & 98.77 \\
        STANet \cite{chen2020spatial}               & 83.81 & \green{91.00} & 87.26 & 77.40 & 98.66 \\
        IFNet \cite{zhang2020deeply}                & \green{94.02} & 82.93 & 88.13 & 78.77 & 98.87 \\
        SNUNet \cite{fang2019dual}                  & 89.18 & 87.17 & 88.16 & 78.83 & 98.82 \\
        BIT \cite{chen2021remote}                   & 89.24 & \blue{89.37}  & \blue{89.31} & \blue{80.68} & \blue{98.92} \\
        Changeformer \cite{bandara2022transformer}  & \blue{92.05} & 88.80  & \red{90.40} & \red{82.48} & \red{99.04} \\
        \hline
        Ours & \red{92.68} & \red{89.47} & \green{91.05} & \green{83.57} & \green{99.10} \\
        \hline
    \end{tabular}
    \label{tab:metrics-levir}
\end{table}

\begin{table}[ht]
    \caption{
        Performance metrics on the WHU-CD dataset. 
        To improve results readability, we adopted a color ranking convention to represent the \green{First}, \red{Second}, and \blue{Third} results respectively.
        The metrics are reported in percentage.
    }
    \centering
    \begin{tabular}{l|ccccc}
        \multicolumn{6}{c}{WHU-CD} \\
        \hline
        Model & Pr & Rc & F1 & IoU & OA \\
        \hline
        FC-EF \cite{daudt2018fully}         & 71.63 & 67.25 & 69.37 & 53.11 & 97.61 \\
        FC-Siam-diff \cite{daudt2018fully}  & 47.33 & 77.66 & 58.81 & 41.66 & 95.63 \\
        FC-Siam-conc \cite{daudt2018fully}  & 60.88 & 73.58 & 66.63 & 49.95 & 97.04 \\
        DTCDSCN \cite{liu2020building}      & 63.92 & \blue{82.30}  & 71.95 & 56.19 & 97.42 \\
        STANet \cite{chen2020spatial}       & 79.37 & \red{85.50 }  & 82.32 & 69.95 & 98.52 \\
        IFNet \cite{zhang2020deeply}        & \green{96.91} & 73.19 & 83.40 & 71.52 & \red{98.83} \\
        SNUNet \cite{fang2019dual}          & 85.60 & 81.49 & \blue{83.50} & \blue{71.67} & 98.71 \\
        BIT \cite{chen2021remote}           & \blue{86.64} & 81.48 & \red{83.98} & \red{72.39 } & \blue{98.75} \\
        \hline
        Ours & \red{91.72} & \green{91.76} & \green{91.74} & \green{84.74} & \green{99.34} \\
        \hline
    \end{tabular}
    \label{tab:metrics-whu}
\end{table}

The baseline models FC-Siam-diff and FC-Siam-conc \cite{daudt2018fully} are the architectures most similar to ours. 
With respect to these two baseline models, we increased the F1 score by 4.73 points on LEVIR-CD, and by more than 20 points on the WHU-CD.
With respect to the best model we found in the literature \cite{bandara2022transformer}, our performance increment on the LEVIR-CD dataset is more limited.
However, as we can see from \tabref{tab:parameters}, our model is 146.50 times smaller.

\begin{table}[ht]
    \caption{
        Parameters, complexity, and performance comparison. 
        The metrics are reported in percentage, parameters in Millions (M), and complexity in GFLOPs (G).
    }
    \centering
    \begin{tabular}{l|c|c|c|c|c}
        \hline
        Model & Param (M) & \thead{Param\\ ratio} & FLOPs (G) &\thead{LEVIR-CD\\F1} &\thead{WHU-CD\\F1} \\
        \hline
        DTCDSCN \cite{liu2020building}& 41.07 & 146.67 & 7.21 &  87.67 & 71.95 \\
        STANet \cite{chen2020spatial}& 16.93 & 60.46 & 6.58 &  87.26 &  82.32  \\
        IFNet \cite{zhang2020deeply}& 50.71 & 181.10 & 41.18 &  88.13 &  83.40 \\
        SNUNet \cite{fang2019dual}& 12.03 & 42.96 & 27.44 &  88.16 &  83.50  \\
        BIT \cite{chen2021remote}& 3.55 & 12.67 & 4.35 &  89.31 &  83.98 \\
        Changeformer \cite{bandara2022transformer}& 41.02 & 146.50 & N.D. &  90.40 & N.D. \\
        \hline
        Ours & 0.28 & 1 & 1.45 & 91.05 & 91.74 \\
        \hline
    \end{tabular}
    \label{tab:parameters}
    \end{table}

In view of these results, we can conclude that our model, despite the lower complexity and the lower number of employed parameters, 
is very effective on the buildings CD task.
Moreover, having not used any global attention mechanism, we have a confirmation of our intuitions: 
in the faced CD task, low level information is sufficient to reach high-quality results. 
Also, the information contained in each single pixel at different resolutions, is very rich and can be exploited to effectively classify changes.

In \figref{fig:bitvsours} a visual/qualitative comparison between the masks created by our model, and those created by BIT 
\cite{chen2021remote} on the LEVIR-CD test dataset, is reported.
Generally speaking, both models perform well and we end up our analysis by conjecturing that the performance difference reported in 
\tabref{tab:metrics-levir} and \tabref{tab:metrics-whu} are more related to missing or hallucinated change regions, than region quality issues.
Nevertheless, we can find some examples were there are significant differences between the ground truth masks (GT) and those created by the two models.
In \figref{fig:bitvsours} it is interesting to note that there are examples where both models fail similarly in the same regions,
despite the two models being based on very different approaches (local versus global).

\ffigure{fig:bitvsours}{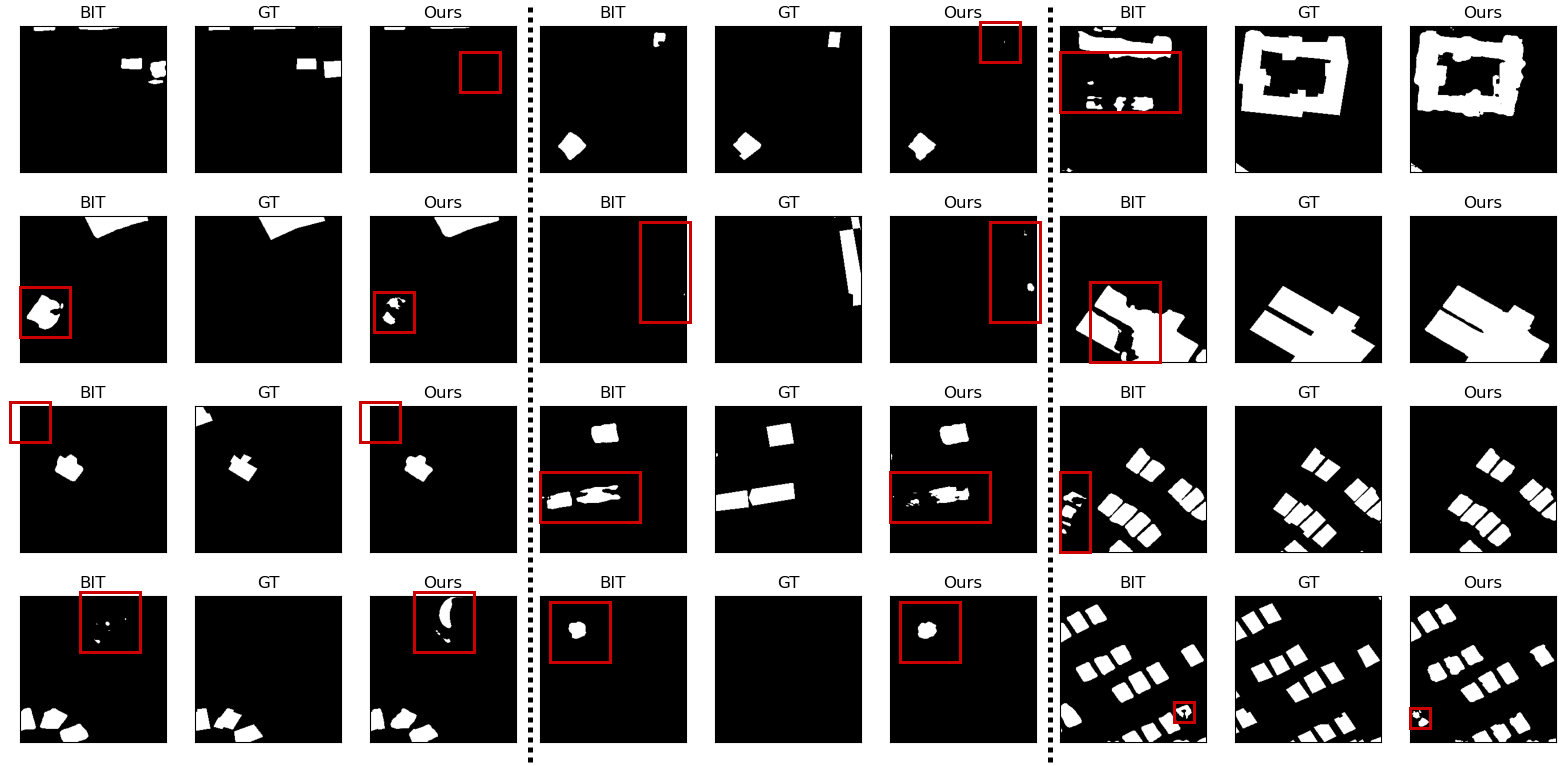}{
    Visual comparison between outputs obtained by our model and BIT. 
    We highlighted with red bounding boxes those regions containing significant differences
    between the ground-truth and the generated masks.
}{10cm}{12cm}

\subsection{Ablation study}
\label{ch:ablation}

In this section we describe the adopted ablation study steps and the achieved results.

\subsubsection{Backbone dimension and final PW-MLP}

The first ablation study we conducted concerns the size of the backbone and the use of the final MLP. 
Regarding the backbone size, we considered both the whole EfficientNet-b4 except the final classifier, 
and a sliced version of the EfficientNet-b4 network including just the first 3 blocks.
Moreover, to assess the effectiveness of the final classification PW-MLP block, 
we considered both the architecture including it, 
and the one that produces its output directly from the last up-sampling block by forcing it output just one channel%
\footnote{
    We employed the sigmoid activation on this output.
}.
The results shown in \tabref{tab:backbonemlp} confirm our intuition on low-level features.
In fact, our solution with the sliced backbone and final PW-MLP, turns out to be the one with the best performances on both datasets.
Furthermore, we note that, to get the best performances, the backbone slicing and PW-MLP classifier must be coupled.
In fact, on LEVIR-CD the backbone slicing only model shows poor performances, 
while the use of the PW-MLP classifier helps the full backbone architecture to improve the quality of the segmentations.
In contrast, on the WHU-CD the architecture with sliced backbone and the PW-MLP classifier obtains better scores 
than the one with full backbone but without PW-MLP, remaining the performances of the latter still unsatisfactory and far from those obtained by our model.

\begin{table}[!ht]
    \caption{
        Performance comparison between versions of our model including and excluding the backbone slicing and the PW-MLP classifier.
    }
    \centering
    \begin{tabular}{l|ccccc|c}
        \hline
        \multicolumn{6}{c}{LEVIR-CD}\\
    \hline
        Model & Precision & Recall & F1 score & IoU & Accuracy & Params \\ 
        \hline
        Full w/o MLP & 83.05 & 94.00 & 88.19 & 78.88 & 98.71 & 17740598 \\ 
        Full w MLP & 92.65 & 89.26 & 90.92 & 83.36 & 99.09 & 17743288 \\ 
        Sliced w/o MLP & 46.15 & 94.52 & 62.02 & 44.95 & 94.10 & 282438 \\ 
        Sliced w MLP & 92.68 & 89.47 & 91.05 & 83.57 & 99.10 & 285128 \\ 
    \hline
        \multicolumn{6}{c}{WHU-CD}\\
        \hline
        Model & Precision & Recall & F1 score & IoU & Accuracy & Params \\ 
        \hline
        Full w/o MLP & 43.08 & 88.12 & 57.87 & 40.72 & 94.91 & 17740598 \\ 
        Full w MLP & 91.00 & 92.14 & 91.57 & 84.45 & 99.32 & 17743288 \\ 
        Sliced w/o MLP & 76.16 & 89.05 & 82.10 & 69.64 & 98.84 & 282438 \\ 
        Sliced w MLP & 91.72 & 91.76 & 91.74 & 84.74 & 99.34 & 285128 \\ 
        \hline
    \end{tabular}
    \label{tab:backbonemlp}
\end{table}

\subsubsection{Impact of skip connection with MAMB}

To quantitatively confirm the usefulness of the skip connections, we trained a model without them 
and compared the achieved results in \tabref{tab:skipnoskip}. 

\begin{table}[ht]
    \caption{Performance comparison between the model with/without skip connections on both datasets LEVIR-CD and WHU-CD.}
    \centering
    \begin{tabular}{l|ccccc}
        \hline
        \multicolumn{6}{c}{LEVIR-CD}\\
        \hline 
        Model type & Pr & Rc & F1 & IoU & OA \\
        \hline
        No Skip & 92.35 & 88.50 & 90.38 & 82.45 & 99.04\\
        Skip & 92.68 & 89.47 & 91.05 & 83.57 & 99.10 \\
        \hline
        \multicolumn{6}{c}{WHU-CD}\\
        \hline
        Model type & Pr & Rc & F1 & IoU & OA \\
        \hline
        No Skip & 90.56 & 89.77 & 90.16 & 82.09 & 99.22\\
        Skip & 91.72 & 91.76 & 91.74 & 84.74 & 99.34 \\
        \hline
    \end{tabular}
    \label{tab:skipnoskip}
\end{table}

As can be seen, all the metrics confirm the beneficial effects of skip connections in the model.
In \figref{fig:maskvis} we reported an example of the intermediate masks that our model creates in the skip-connections.

\ffigure{fig:maskvis}{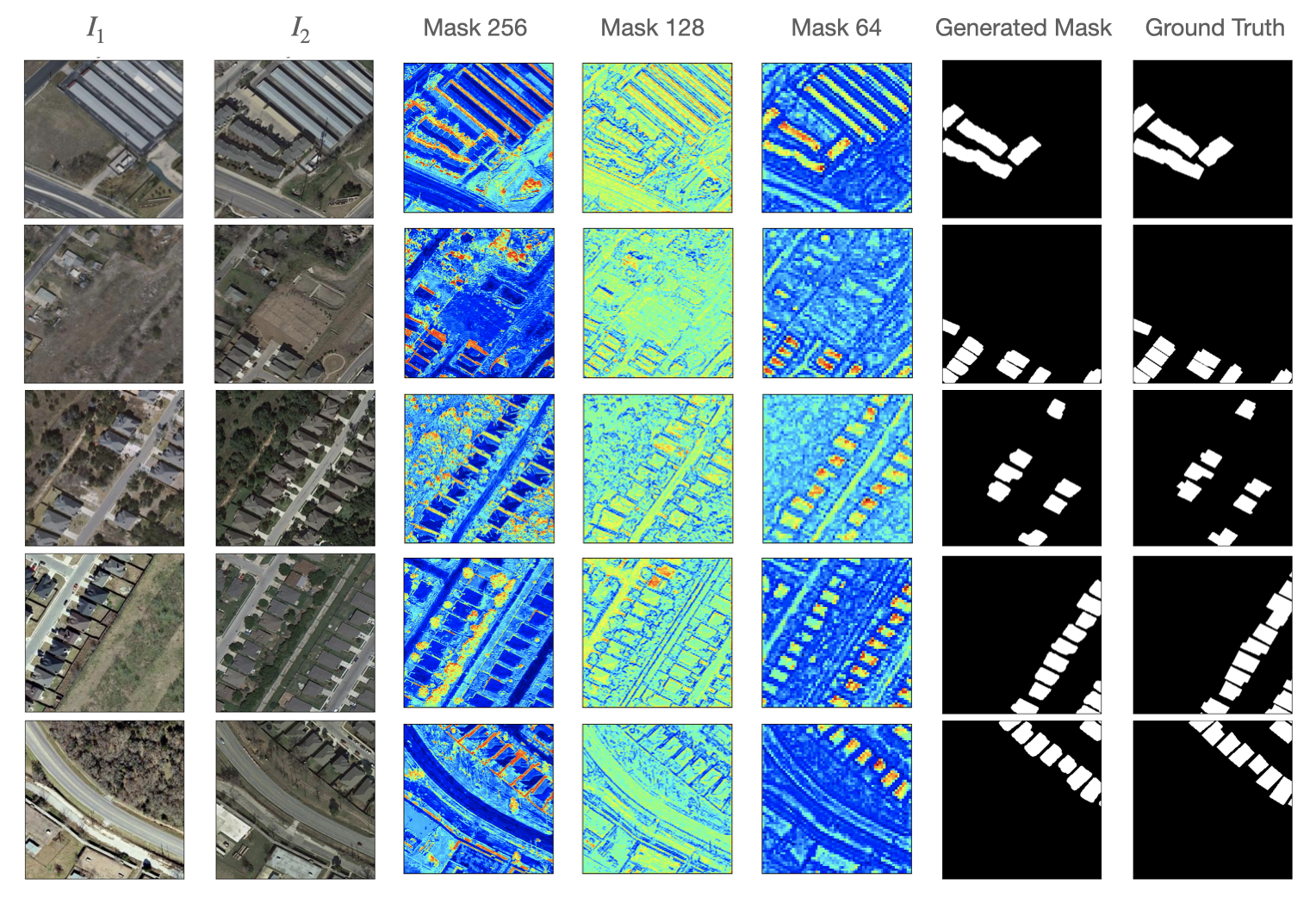}{
    Visualization of the intermediate masks at different resolutions and the final binary mask for one example image pair.
}{10cm}{12cm}

As can be seen, the masks created with the MAMB block at resolution 64 highlight the objects that musk be tracked (red pixels).
The intermediate masks at resolution 128 act more like an edge detector.
Finally, the masks at resolution 256, obtained applying the MAMB block directly to the original images $I_1$ and $I_2$, 
distinguish between object classes like buildings and street (dark blue), vegetation (light green), and shadows (red). 
The ability to highlight shadows is very effective since it helps the model to detect objects and to refine their edges.

\subsubsection{Comparison with other simple mixing strategy}

In \tabref{tab:groupedvsconv} we compare our mixing strategy, described in \secref{sec:mixing}, 
with other widely used feature fusion blocks. We tested the following alternatives:
\begin{itemize}
    \item subtraction, both in the bottleneck and in skip connections;
    \item concatenation + convolution, both in the bottleneck and in skip connections.
\end{itemize}
We selected these two alternatives since our mixing strategy can be seen as a generalization of the pixel-wise subtraction%
\footnote{
    In fact, if we initialize all of our 2-depth kernels with the "central" weights to $1$ and $-1$, 
    and all the rest to $0$, we have the standard subtraction.
}.
However, our mixing block \secref{sec:mixing} is fully trainable with the spirit of feature re-use \cite{bengio2013representation}.
Moreover, concatenation + convolution can be seen as generalization of our mixing block. 
However, the number of trainable parameters to be tuned for this mixing block is much bigger than ours.
More precisely, the number of parameters in our mixing block is $c(2 \cdot k_h \cdot k_w)$,
\footnote{
    The parentheses are highlighting the size of each kernel and the number of kernels.
}
where $c$ is the number of channels, $k_h$, $k_w$ are the convolutional kernel sizes.
By comparison, a convolution working on the concatenated feature tensors contains $c(2c \cdot k_h \cdot k_w)$ parameters.
 
\begin{table}[!ht]
    \caption{
        Performance comparison between the model with our mixing strategy, subtraction, and concatenation + convolution (C+C) respectively.
    }
    \centering
    \begin{tabular}{l|ccccc|cc}
    \hline 
    \multicolumn{8}{c}{LEVIR-CD} \\
    \hline
        Model type & Pr & Rc & F1 & IoU & OA & Param. tot. & GFLOPs $\pm$ \\
        \hline 
        Subtraction & 92.13 & 89.41 & 90.75 & 83.07 & 99.07 & 282939 & 1.43 ($-1.4\%$) \\ 
        C+C & 92.55 & 89.61 & 91.06 & 83.59 & 99.10 & 368468 & 1.75 ($+20.7\%$) \\
        \hline
        Our & 92.68 & 89.47 & 91.05 & 83.57 & 99.10 & 285128 & 1.45 \\
        \hline
        \multicolumn{8}{c}{WHU-CD} \\
        \hline
        Model type & Pr & Rc & F1 & IoU & OA & Param. tot. & GFLOPs $\pm$ \\
        \hline
        Subtraction & 90.10 & 91.55 & 90.82 & 83.19 & 99.26 & 282939 & 1.43 ($-1.4\%$) \\
        C+C & 92.19 & 91.25 & 91.72 & 84.71 & 99.34 & 368468 & 1.75 ($+20.7\%$) \\
        \hline
        Our & 91.72 & 91.76 & 91.74 & 84.74 & 99.34 & 285128 & 1.45 \\
        \hline
    \end{tabular}
    \label{tab:groupedvsconv}
\end{table}

In \tabref{tab:gridsearchmix} we reported the results of a more detailed study on mixing strategies.
We alternated the use of subtraction/concatenation + convolution with our respective proposal 
to mix the features in the bottleneck/skip connections.

\begin{table}[!htb]
    \caption{
        Evaluation of subtraction and concatenation + convolution mixing strategies.
        We reported F1 score for the two datasets LEVIR-CD (F1-L) and WHU-CD (F1-W).
        We used \cmark to indicate where we changed our proposed option with subtraction or concatenation + convolution.
        In contrast, \xmark represents our bottleneck mixing block or MAMB.
    }
    \label{tab:gridsearchmix}
    \begin{subtable}{.5\linewidth}
      \centering
        \caption{Subtraction}
        \begin{tabular}{c|c|c|c|c}
        Mix & Skip &  F1-L & F1-W & Param\\
        \hline
            \cmark & \cmark &  90.75 & 90.82 & 282939 \\
            \xmark & \cmark &  90.75 & 91.51 & 284004 \\  
            \cmark & \xmark &  90.71 & 89.58 & 284063 \\
            \hline
            \xmark & \xmark &  91.05 & 91.74 & 285128 \\
            \hline
        \end{tabular}
    \end{subtable}%
    \begin{subtable}{.5\linewidth}
      \centering
        \caption{Concatenation+Convolution}
        \begin{tabular}{c|c|c|c|c}
        Mix & Skip &  F1-L & F1-W & Param \\
        \hline
            \cmark & \cmark &  91.06 & 91.72 & 368468 \\ 
            \xmark & \cmark &  91.06 & 91.08 & 313028 \\ 
            \cmark & \xmark &  90.90 & 91.71 & 340568 \\
            \hline
            \xmark & \xmark &  91.05 & 91.74 & 285128 \\
            \hline
        \end{tabular}
    \end{subtable} 
\end{table}

The obtained results confirm that our proposal can be considered an effective generalization of the subtraction, 
with little impact on the size and complexity of the model.
On the other hand, the overhead introduced by the concatenation + convolution mixing strategy,
seems to produce little differences in terms of performance.

\subsubsection{Channel-wise MLP vs CycleMLP}

As reported in \secref{chp:relationexisting}, several MLP blocks have recently been studied with the intent of 
incorporating both spatial and channel-specific information.
As previously described, we used the MLPs only along the channels in the final classifier, 
and coupled to our mixing strategy in the MAMB blocks to obtain space-time correlation.
We then decided to deal with the CycleMLP block proposed in \cite{chen2021cyclemlp}.
The results reported in \tabref{tab:mlpvscyclemlp} suggest the superiority of our proposed
use of MLPs compared to that proposed in \cite{chen2021cyclemlp}.
A heuristic explanation for these results can be the following:
the MLP blocks proposed in \cite{chen2021cyclemlp} have shown to obtain excellent performances 
when they are used to construct a hierarchical architecture to generate pyramid features.
This makes us think that the advantage of CycleMLPs may be more significant 
when the features are more refined than the low-level features we use.

\begin{table}[!ht]
    \caption{
        Performance comparison between MLP and CycleMLP \cite{chen2021cyclemlp} on LEVIR-CD and WHU-CD.
        We used \cmark to indicate experiments where we changed our proposed block with a CycleMLP one,
        while \xmark represents our proposed architecture.
    }
    \centering
    \begin{tabular}{c|c|ccccc|c}
    \hline 
    \multicolumn{7}{c}{LEVIR-CD} \\
    \hline
        Skip & Class. & Pr & Rc & F1 & IoU & OA & Param. tot. \\
        \hline 
        \cmark & \xmark & 92.47 & 88.48 & 90.43 & 82.53 & 99.04 & 309300 \\ 
        \cmark & \cmark & 92.45 & 88.49 & 90.42 & 82.52 & 99.04 & 314542 \\ 
        \xmark & \cmark & 92.58 & 88.96 & 90.73 & 83.04 & 99.07 & 290370 \\ 
        \hline
        \xmark & \xmark & 92.68 & 89.47 & 91.05 & 83.57 & 99.10 & 285128 \\ 
        \hline
        \multicolumn{7}{c}{WHU-CD} \\
        \hline
        Skip & Class. & Pr & Rc & F1 & IoU & OA & Param. tot. \\
        \hline
        \cmark & \xmark & 89.76 & 89.06 & 89.41 & 80.85 & 99.16 & 309300 \\ 
        \cmark & \cmark & 92.25 & 90.51 & 91.37 & 84.12 & 99.32 & 314542 \\ 
        \xmark & \cmark & 90.20 & 85.84 & 87.96 & 78.52 & 99.06 & 290370 \\ 
        \hline
        \xmark & \xmark & 91.72 & 91.76 & 91.74 & 84.74 & 99.34 & 285128 \\ 
        \hline
    \end{tabular}
    \label{tab:mlpvscyclemlp}
\end{table}

\section{Limitations of our work} \label{chp:limitations}

In all the experiments we performed, we observed our model learning some domain-specific patterns.
Despite being this an advantage allowing the model to better deal with the faced task,
this is also a limitation because it reduces the model's ability to adapt to new scenarios by means of fine-tuning.
Furthermore, in the two datasets taken into consideration, and also in our industrial case study, 
the images $I_1$ and $I_2$ are spatially registered. 
This allowed the successful usage of low-level features without assessing to global feature relationships.
In different contexts, where the images undergo large spatial shifts, 
this local approach can show worse performances with respect to more global approaches 
like vision transformers \cite{chen2021remote,bandara2022transformer}.

\section{Conclusions and future works} \label{cap:conclusion}

Guided by industrial needs, we proposed a tiny convolutional change-detection Siamese U-Net like model.

Our model exploits low-level features by comparing and classifying them to obtain a binary map of detected changes.
We propose a mixing block showing the ability to compare/compose features on both spatial and temporal domains.
The proposed PW-MLP block shows great ability in extracting features useful to classify occurred changes on a per-pixel basis.
The composition of these proposed blocks, here referred as MAMB, 
shows the ability to estimate masks useful to enrich features used in the U-Net decoder part.
We have shown that an effective way to generate the output mask is to process low-level backbone features in a PW-MLP block,
effectively facing the change-detection task as a per-pixel classification problem.

We tested our model on public change-detection datasets containing aerial images acquired at two different times.
Furthermore, we compared the achieved results with SOTA models proposed in the change-detection literature.
Our tests demonstrated that our model performs comparably or better than the current SOTA models,
remaining at the same time the smaller and faster one.

Notice that the ideas employed in this work can be also applied to other fields. 
For this reason, we will investigate the application of MAMB and PW-MLP blocks 
to tasks such as anomaly-detection, surveillance, and semantic segmentation.

As a future work, solutions to the limitations presented in \ref{chp:limitations} will be investigated.
Moreover, in order to be able to extend our approach even in those contexts where global features play a fundamental role,
we would like to explore multibranch models in which one branch works on local features and one on global features.

\section*{Acknowledgments}
The authors want to thank the whole ARGO Vision team, professor Stefano Gualandi, Gabriele Loli and Gennaro Auricchio for the useful discussion and comments. 
We also want to thank all those who have provided their codes in an accessible and reproducible way.
The Ph.D. scholarship of Andrea Codegoni is founded by SeaVision s.r.l..
\newpage
\appendix

\section{Backbones comparison}
\label{appx:backbone}

In this appendix we report the results obtained by varying the backbone adopted in the model.
In each backbone we select only few initial blocks in order to work with features
that are not very complex and not excessively aggregated from a spatial point of view.
We decide to select the all the initial blocks up to the first having spatial resolution 32x32.
Due to the different compositions of the considered networks, 
the final size of the models changes it the range starting from a minimum of 32k parameters up to 1.3M.

As we can see from \tabref{tab:backbones}, the results obtained are stable from the performances point of view.
The backbones of the EfficientNet family appear to be, in accordance with the experiments carried out on our proprietary dataset, 
those that achieve the best performances.
However, the other backbone types also produce comparable results making our approach:
\begin{itemize}
    \item robust with respect to the backbone used;
    \item flexible with respect to the required size and computational complexity.
\end{itemize}

In this comparison we have not considered Transformers-type backbones such as \cite{liu2021swin,dosovitskiy2020image}.
The reason for this choice lies in the fact that the philosophy of the Transformers is a global philosophy, 
as opposed to the blocks we propose which are instead local.
As mentioned in \secref{cap:conclusion}, an integration of these two philosophies will be the subject of future works.

\begin{table}[!ht]
    \centering
    \caption{Comparison of different backbones on LEVIR-CD dataset}
    \begin{tabular}{l|ccccc|c}
    \hline 
    \multicolumn{7}{c}{LEVIR-CD} \\
    \hline
        Backbone & Precision & Recall & F1 score & IoU & Accuracy & Params \\ 
        \hline
        mobilenetv2 & 90.95 & 86.43 & 88.63 & 79.59 & 98.87 & 38798 \\ 
        mobilenetv3large & 90.56 & 85.98 & 88.21 & 78.91 & 98.82 & 32886 \\ 
        resnet18 & 92.15 & 87.43 & 89.72 & 81.37 & 98.98 & 707894 \\ 
        efficientnetb0 & 92.18 & 87.96 & 90.02 & 81.85 & 99.00 & 79480 \\ 
        efficientnetb1 & 92.17 & 88.92 & 90.51 & 82.67 & 99.05 & 122092 \\ 
        efficientnetb2 & 92.13 & 89.26 & 90.68 & 82.94 & 99.06 & 148040 \\ 
        efficientnetb3 & 92.40 & 89.54 & 90.95 & 83.40 & 99.09 & 178716 \\ 
        efficientnetb4 & 92.68 & 89.47 & 91.05 & 83.57 & 99.10 & 285128 \\ 
        mnasnet13 & 91.95 & 88.17 & 90.02 & 81.86 & 99.00 & 97262 \\
        densenet121 & 92.13 & 87.97 & 90.00 & 81.83 & 99.00 & 1364790 \\
        \hline
    \end{tabular}
    \label{tab:backbones}
\end{table}

\section{Hyperparameters' tuning}
\label{appx:A1}

In this appendix we report the details about the hyperparameters' tuning experiments.
One of the advantages of using limited computational complexity models is 
being able to fine-tune hyperparameters using relatively few computational resources 
and in a reasonable time from an industrial point of view.
In our experiments we tune the learning rate, the weight decay, and the usage of the \emph{amsgrad} strategy.
The framework used to run the experiments and optimize the hyperparameters is NNI \cite{nni2021}.

Since we execute only 100 epochs per run, we chose a higher learning rate range $(10^{-3},4\cdot 10^{-3})$,
in order to explore whether a higher than standard learning rate leads to faster model convergence.
As for the weight decay, we follow a conservative choice by setting the range between $10^{-2}$ and $8\cdot 10^{-3}$.
We also test other simple loss functions for model training such as Mean Square Error (MSE), 
Intersection over Union (IoU) and a combination of IoU and BCE.

In \figref{fig:hyper} we show the various combinations of hyperparameters explored in a batch of 30 experiments,
and the relative performances on the LEVIR-CD validation set.
Analyzing the results, we note that BCE and MSE, 
regardless of the other parameters, obtain superior performance compared to the IoU.
In addition, the BCE + IoU combination, although better than IoU, also scores lower than the BCE and MSE.
Regarding the other hyperparameters, as can be seen in particular from \figref{fig:metricstab}, 
our model obtains robust performances with respect to all the tested combinations.
Finally, we note that in the conducted experiments, 
BCE has lower variance in terms of F1 score with respect to the choices of the other hyperparameters.
This represents another motivation for us to chose BCE as loss function. 

\ffigure{fig:hyper}{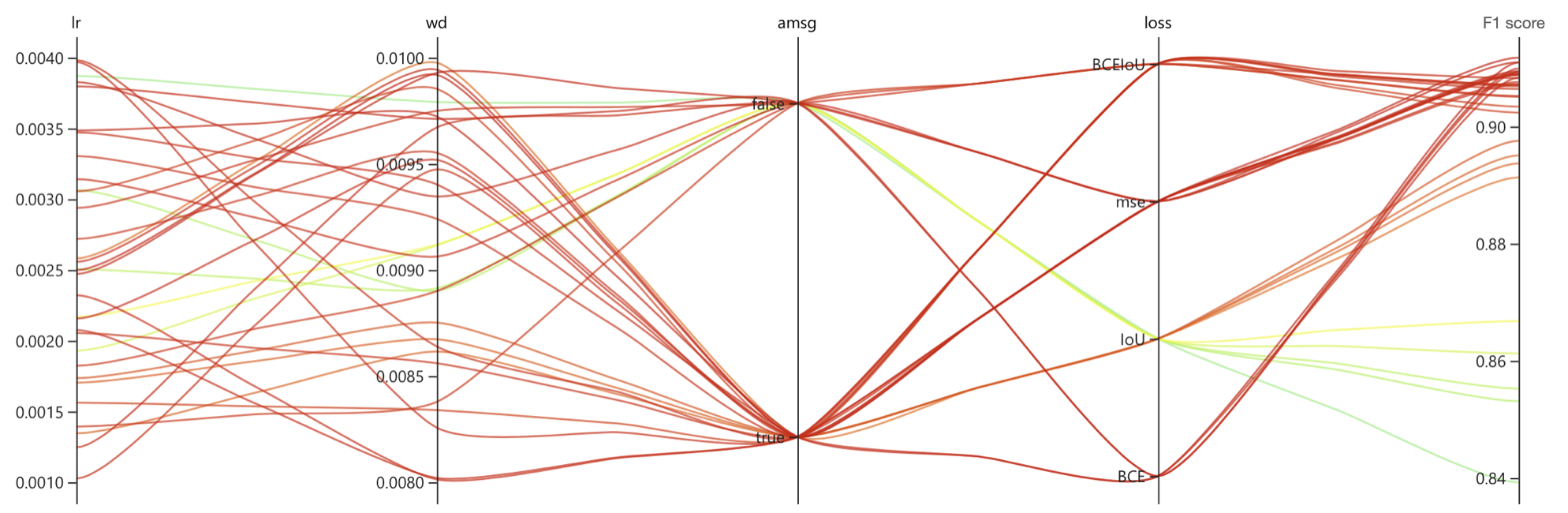}{
    Different combination of parameters and their impact on the F1 score on the LEVIR-CD dataset.
}{3cm}{10cm}

\ffigure{fig:metricstab}{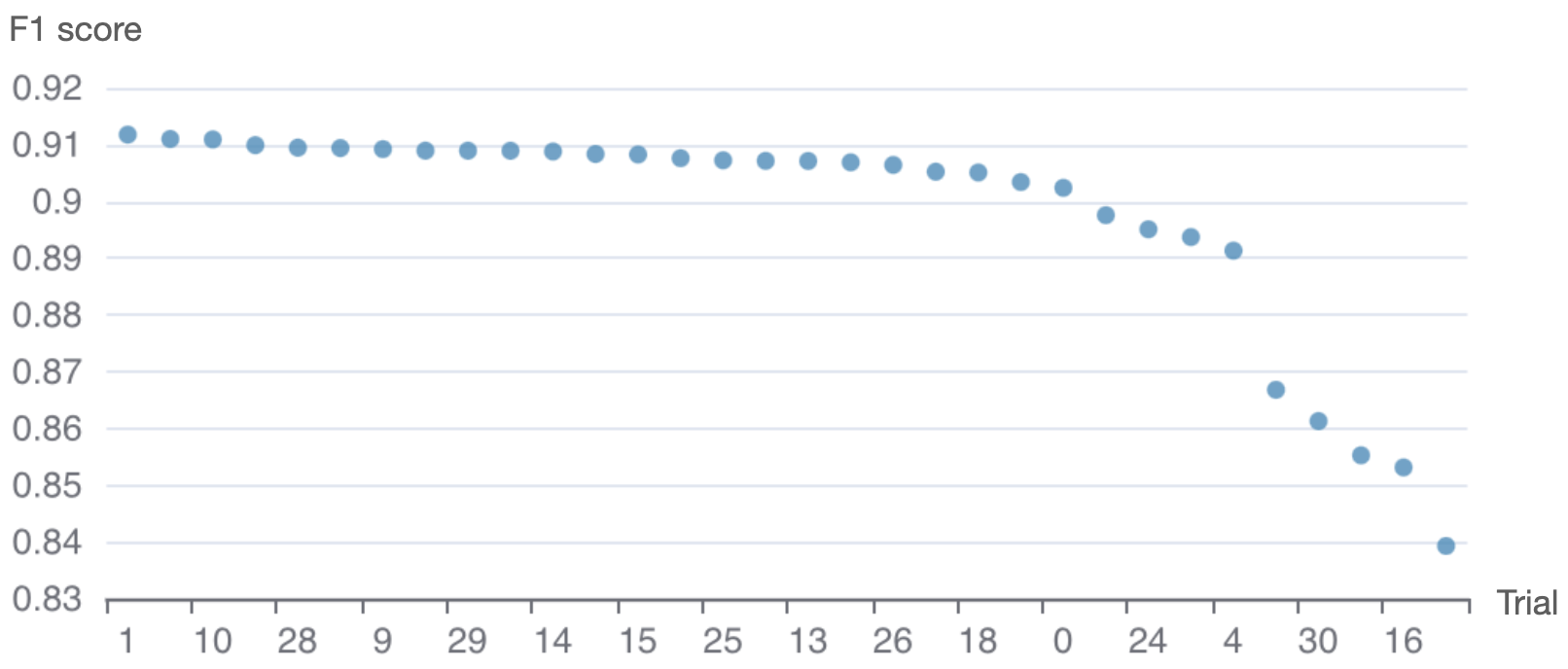}{
    Behavior of the final F1 score in the different experiments conducted to tune the hyperparameters.
    The drop in the F1 score is due to the use of IoU as loss function.
}{3cm}{10cm}

\bibliographystyle{IEEEtran}  
\bibliography{references}

\end{document}